%% file: main.tex
\documentclass[10pt,twocolumn,letterpaper]{article}

\usepackage[pagenumbers]{cvpr} 


\definecolor{cvprblue}{rgb}{0.21,0.49,0.74}
\usepackage[pagebackref,breaklinks,colorlinks,allcolors=cvprblue]{hyperref}

\def\paperID{12658} 
\def\confName{CVPR}
\def\confYear{2026}

\title{IMTalker: Efficient Audio-driven Talking Face Generation with Implicit Motion Transfer}

\author{
    Bo Chen$^{1,2}$ \quad 
    Tao Liu$^{1}$ \quad 
    Qi Chen$^{1}$ \quad 
    Xie Chen$^{1,*}$ \quad 
    Zilong Zheng$^{2,*}$ \\[5pt] 
    $^{1}$X-LANCE Lab, Shanghai Jiao Tong University \quad \\
    $^{2}$State Key Laboratory of General Artificial Intelligence, BIGAI \\
    {\tt\small \{cbforever, chenxie95\}@sjtu.edu.cn} \quad 
    {\tt\small zlzheng@bigai.ai} \\[3pt] 
    \small \textbf{Project:} \url{https://cbsjtu01.github.io/IMTalker/}
}
\usepackage{siunitx}
\usepackage{multirow}
\usepackage{booktabs}
\usepackage{adjustbox}
\usepackage{caption}
\begin{document}

\setlength{\textfloatsep}{10pt plus 1pt minus 2pt}

\setlength{\floatsep}{10pt plus 1pt minus 2pt}

\setlength{\intextsep}{5pt plus 1pt minus 2pt}

\setlength{\abovecaptionskip}{5pt}
\setlength{\belowcaptionskip}{0pt}
\input{sec/teaser}
\let\thefootnote\relax\footnotetext{* Corresponding authors.}
\input{sec/0_abstract}    
\input{sec/1_Introduction}
\input{sec/2_Related_Work}
\input{sec/3_Method}
\input{sec/4_Experiment}

\input{sec/5_Conclusion}

{
    \small
    \bibliographystyle{ieeenat_fullname}
    \bibliography{main}
}


\end{document}

%% file: sec/teaser.tex
\twocolumn[{
\renewcommand\twocolumn[1][]{#1}
\maketitle
\begin{center}
    \centering
    \vspace{-5pt}
    \includegraphics[width=\linewidth]{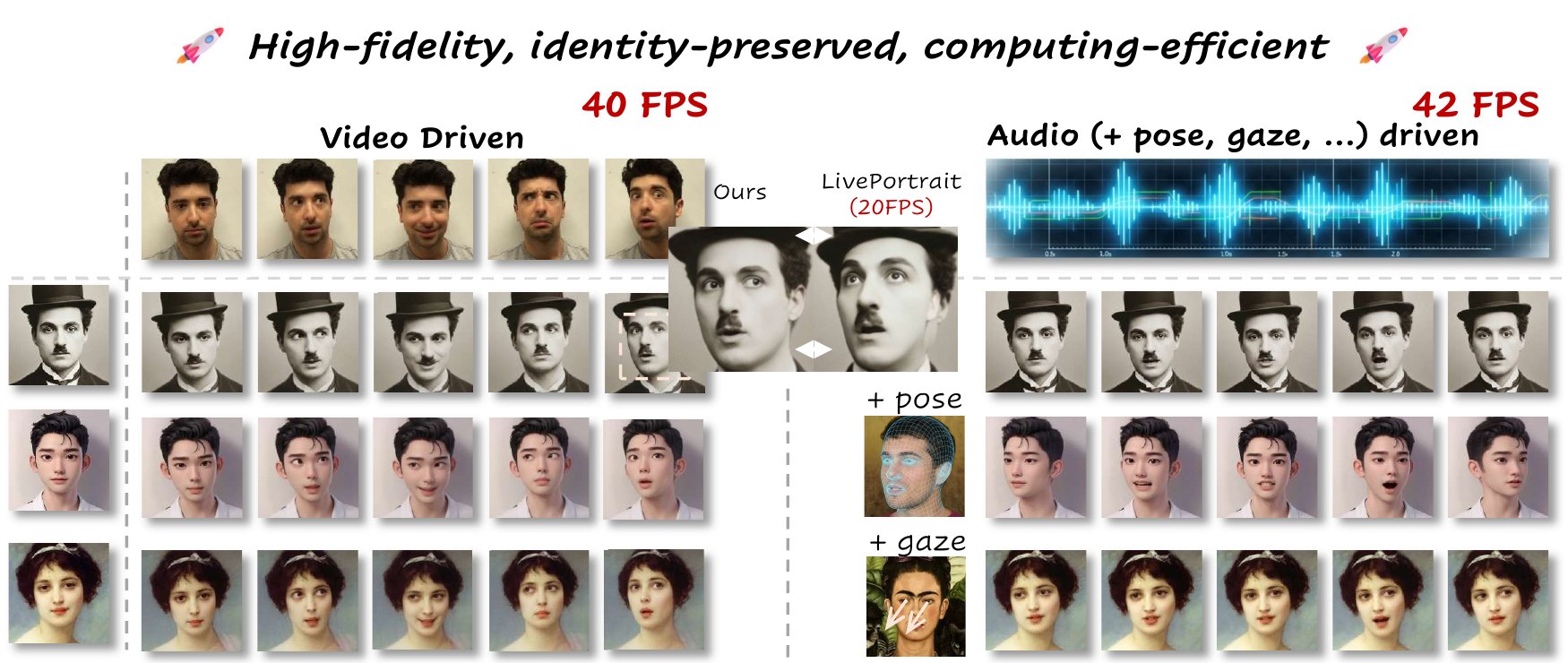}
    \setlength{\abovecaptionskip}{0mm}
    \vspace{-5pt}
    \captionof{figure}
    {IMTalker accepts diverse portrait styles and achieves 40 FPS for video-driven and 42 FPS for audio-driven talking-face generation when tested on an NVIDIA RTX 4090 GPU at $512\times512$ resolution. It also enables diverse controllability by allowing precise head-pose and eye-gaze inputs alongside audio.}
	\label{fig:teaser}
\end{center}
}]

%% file: sec/0_abstract.tex
\begin{abstract}


Talking face generation aims to synthesize realistic speaking portraits from a single image, yet existing methods often rely on explicit optical flow and local warping, which fail to model complex global motions and cause identity drift. We present ~\emph{IMTalker}, a novel framework that achieves efficient and high-fidelity talking face generation through implicit motion transfer. The core idea is to replace traditional flow-based warping with a cross-attention mechanism that implicitly models motion discrepancy and identity alignment within a unified latent space, enabling robust global motion rendering. To further preserve speaker identity during cross-identity reenactment, we introduce an identity-adaptive module that projects motion latents into personalized spaces, ensuring clear disentanglement between motion and identity. In addition, a lightweight flow-matching motion generator produces vivid and controllable implicit motion vectors from audio, pose, and gaze cues. Extensive experiments demonstrate that IMTalker surpasses prior methods in motion accuracy, identity preservation, and audio–lip synchronization, achieving state-of-the-art quality with superior efficiency, operating at 40 FPS for video-driven and 42 FPS for audio-driven generation on an RTX 4090 GPU. We will release our code and pre-trained models to facilitate applications and future research.
\end{abstract}

%% file: sec/1_Introduction.tex
\section{Introduction}
\label{sec:intro}
Talking face generation creates a photorealistic video of a person speaking from a single portrait, guided by either an audio track or a driving video. Owing to its applications in virtual avatars, video conferencing, and digital content creation, it has become a major research focus. An ideal model must satisfy a set of often competing desiderata: high-fidelity generation, precise synchronization with the driving source (such as lip-to-audio or motion-to-video), strong identity preservation, natural facial expressiveness, and real-time performance. The realization of such an ideal model is an indispensable step toward more advanced human-computer interaction.

Driven by the rapid advances of generative models such as GANs \cite{goodfellow2014generative} and diffusion models \cite{ho2020denoising, nichol2021improved, rombach2022high}, talking face generation has achieved remarkable progress. Recent approaches \cite{liu2024anitalker, zhang2023sadtalker, tan2024edtalk, ki2025float, ji2025sonic} can synthesize expressive facial dynamics and natural head movements.
Existing methods generally follow a two-stage pipeline \cite{zhang2023sadtalker, liu2024anitalker, tan2024edtalk, li2024ditto, ki2025float} for efficiency, 
which first converts the audio into an identity-agnostic motion representation and then estimates explicit optical flow to warp the source frame \cite{siarohin2019first, drobyshev2024emoportraits, guo2024liveportrait, wang2022latent, wang2021one}. This explicit, local warping is efficient but inherently assumes small, locally smooth deformations. As a result, it struggles to handle large head poses, occlusions, and non-rigid articulations, often producing stretching, tearing, or identity drift that degrade realism.
These limitations highlight the need for a \textit{global}, \textit{implicit motion representation} capable of modeling complex movements while maintaining identity consistency and efficiency.

To address the above limitations, we introduce \emph{IMTalker}, an efficient and realistic talking-face generation framework built upon an \textbf{I}mplicit \textbf{M}otion representation.
Instead of explicitly predicting pixel-wise optical flow and performing local warping, \emph{IMTalker} encodes all identity-agnostic motion cues into a compact latent embedding and employs a cross-attention–based implicit motion transfer module to align motion discrepancies with identity features in a unified latent space.
This implicit formulation captures global and non-rigid facial dynamics, enabling holistic re-rendering of the face and substantially mitigating the identity degradation that occurs under large-pose variations or expressive movements.

However, simultaneously combining motion transfer and identity alignment within the latent space inevitably introduces an inductive bias related to identity information into the motion latent. During cross-identity motion transfer, this state of incomplete disentanglement between identity and motion challenges the model's ability to maintain identity consistency. To address this, we design an Identity-Adaptive module. Specifically, we project the same motion latent vector into the respective motion spaces of different subjects to derive personalized descriptions of that motion. We then apply a distance constraint to regularize these personalized representations during the motion transfer. Through this adaptive process, we efficiently achieve identity-motion disentanglement, thereby preserving identity consistency during the motion transfer process.


Furthermore, we develop a conditional Flow-Matching Motion Generator to produce vivid and controllable motion latents from audio, pose, and gaze signals. This generator benefits from the stability and efficiency of Flow Matching~\cite{lipman2022flow, liu2022flow}, enabling diverse and realistic facial dynamics under real-time settings.
Extensive experiments on HDTF \cite{zhang2021flow} and CelebV \cite{yu2023celebv} confirm that IMTalker outperforms existing flow-based and diffusion-based approaches in motion accuracy, identity preservation, and audio–lip synchronization.

Our key contributions are summarized as follows:
\begin{itemize}

\item We introduce IMTalker, a new efficient and realistic talking face generation framework. Its core novelty lies in replacing explicit optical flow and warping with attention-based implicit motion transfer, which holistically models global motion and improves realism.

\item We propose a dedicated Identity-Adaptive module to explicitly tackle the problem of identity leakage during motion transfer. This ensures that the generated video maintains a high degree of consistency with the source identity.

\item We design a lightweight and versatile motion generator using a conditional flow matching model. This allows for fine-grained control over facial expression, head pose, and eye movement, leading to more vivid and dynamic video synthesis.

\item Extensive experiments demonstrate that our method achieves state-of-the-art performance in precise motion transfer, high identity preservation, and superior audio-to-lip synchronization.
\end{itemize}

%% file: sec/2_Related_Work.tex
\section{Related Work}
\label{sec:related_work}
\subsection{Audio-Driven talking face}


With the emergence of diffusion models \cite{ho2020denoising,song2020denoising}, several works (e.g., EMO \cite{tian2024emo}, Hallo-1/2/3 \cite{xu2024hallo,cui2024hallo2,cui2025hallo3}, Sonic \cite{ji2025sonic}) have explored end-to-end audio-driven talking face generation by injecting multimodal cues, such as audio, motion reference, and appearance reference—into pretrained video diffusion models through a Reference Network.
Although these methods enable more holistic facial animation compared with earlier lip-synchronization approaches, they often suffer from appearance leakage from the motion reference, leading to visual artifacts and high computational cost.

To address these challenges, subsequent research has increasingly adopted a two-stage framework to explicitly disentangle appearance and motion. In the first stage, they construct a disentangled motion and appearance space; in the second stage, a separate network maps audio features to the motion representation.

Regarding motion representation, some approaches employ explicit representations, such as facial landmarks (e.g., EchoMimic \cite{chen2025echomimic}, SPACEx \cite{gururani2023space}) or 3D morphable model (3DMM) coefficients (e.g., SadTalker \cite{zhang2023sadtalker}). These methods generally achieve strong audio-visual consistency and controllability over the entire face. However, due to the limited expressiveness of landmarks and 3DMM parameters, they often fail to capture subtle and stochastic facial movements, resulting in less natural motion synthesis.

Alternatively, other works learn implicit motion representations through AutoEncoders that disentangle motion and appearance in the latent space. For example, 
AniTalker \cite{liu2024anitalker} enhances disentanglement using speaker identity and mutual information constraints; Ditto \cite{li2024ditto}, inspired by Face-vid2vid \cite{wang2021one} and LivePortrait \cite{guo2024liveportrait}, represents facial motion using learnable facial keypoints; and FLOAT \cite{ki2025float}, following LIA \cite{wang2022latent}, achieves disentanglement through a linear orthogonal structure to compress motion features.

For audio-to-motion generation, most recent methods adopt diffusion models or flow matching \cite{lipman2022flow} for sequence-to-sequence generation, as demonstrated by AniTalker and FLOAT.

\begin{figure*}[t!]
    \centering
    \includegraphics[width=\linewidth]{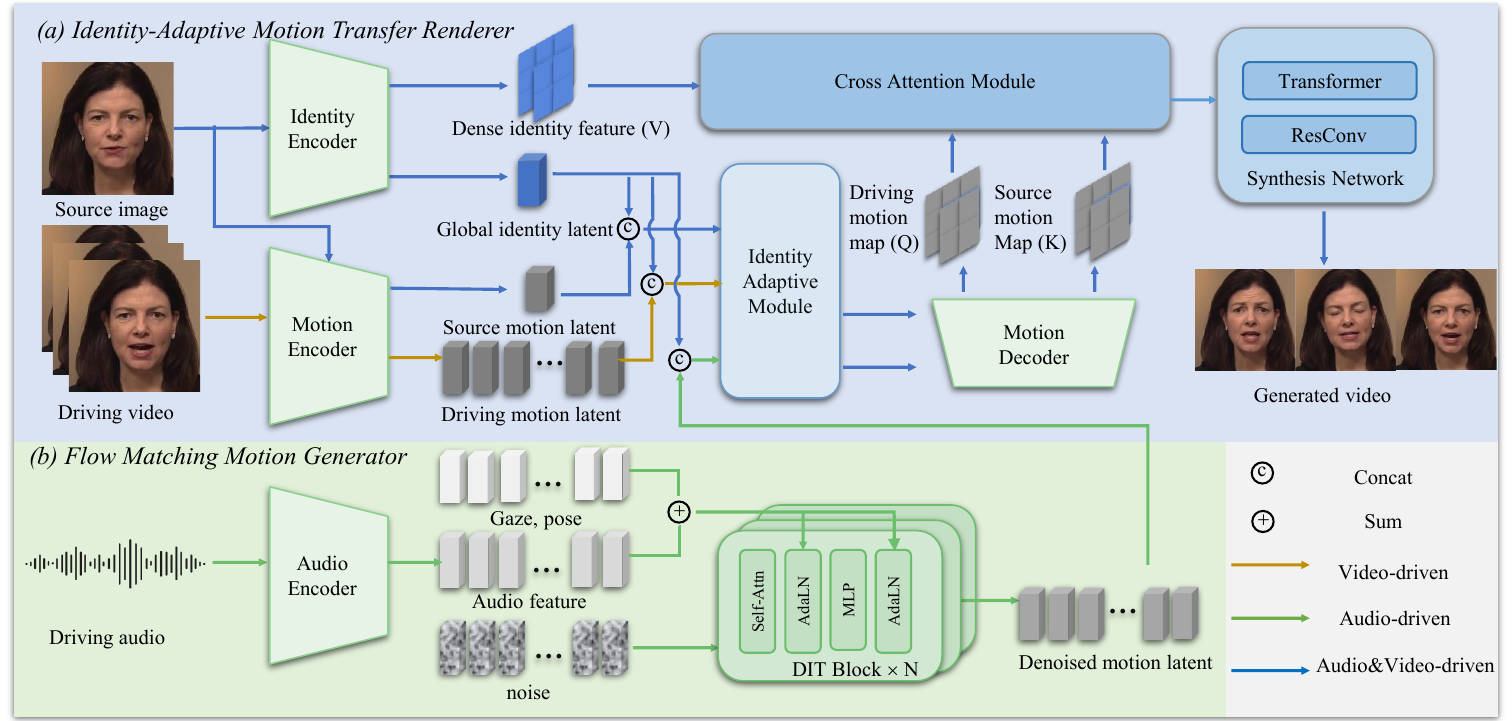}
    \caption{
        \textbf{Overall architecture of our proposed framework.} 
    Given a source image, an \emph{Identity Encoder} and \emph{Motion Encoder} extract identity features and source motion. 
    A driving motion latent is either extracted from video by the \emph{Motion Encoder} or synthesized from audio by a \emph{Flow-Matching Motion Generator}. 
    Both motion latents are then personalized by the \emph{Identity-Adaptive Module}. 
    Subsequently, the \emph{Implicit Motion Transfer  Module} uses these personalized latents and $f_{id}$ to generate aligned features via a motion decoder and cross-attention. 
    Finally, the \emph{Synthesis Network} renders these aligned features into the final photorealistic image.
    }
    \label{fig:pipeline}
    \vspace{-10pt}
\end{figure*}
\subsection{Video-Driven talking face}
To achieve a disentangled representation of motion and appearance, many audio-driven talking face approaches build upon the architectures of video-driven models. Face-vid2vid \cite{wang2021one} employs implicit 3D keypoints to represent facial motion and applies affine transformations for face warping, laying the foundation for subsequent methods. Following this idea, MCNet \cite{hong23implicit} and LivePortrait \cite{guo2024liveportrait} further refine motion representation and improve reconstruction fidelity. X-Portrait \cite{xie2024x}, built upon a pre-trained Face-vid2vid model, enhances attention to local facial movements, thereby mitigating appearance leakage.
In addition, several works adopt orthogonal bases to represent motion, modeling the motion code as a linear combination of a set of basis vectors. For example, LIA \cite{wang2022latent}, LIA-X \cite{wang2025lia}, EDTalk \cite{tan2024edtalk} and FIXTalk \cite{tan2025fixtalk} maintain a set of learnable orthogonal embeddings during training, enabling more stable and disentangled motion encoding.
Moreover, some approaches leverage information compression techniques combined with attention mechanisms to aggregate information from videos and achieve disentanglement. For instance, IMF \cite{gao2024implicit} proposes an implicit motion function that aggregates motion information from videos using low-dimensional vectors and a cross-attention mechanism. Similarly, BCD \cite{li2025bitrate} constrains the bitrate of motion using GVQ, enabling implicit disentanglement of motion representation.

%% file: sec/3_Method.tex
\section{Methodology}
\label{sec:method}
We will first introduce the overview of our framework \cref{sec:overview}, followed by a detailed description of the two main components: (1) an \textbf{Identity-Adaptive Motion Transfer Renderer} \cref{sec:render} responsible for personalizing motion representations, achieving motion transfer in the latent space, and finally synthesizing photorealistic facial imagery, and (2) a \textbf{Flow-Matching Motion Generator} \cref{sec:generator} that learns to generate identity-agnostic facial motion latent vectors from audio and other control conditions.  

\subsection{Framework Overview}
\label{sec:overview}

Let $I_S \in \mathbb{R}^{H \times W \times 3}$ be a static source identity image and $\mathcal{D}$ be a driving signal. Our framework, IMTalker, is designed to handle two distinct reenactment tasks by accepting two types of driving signals: an audio waveform $A$ for audio-driven synthesis, or a driving video sequence $V_D$ for video-driven motion transfer. The objective is to synthesize a video frame sequence $\{\hat{I}_T\}$ that preserves the identity of $I_S$ while exhibiting facial motion driven by the driving signal $\mathcal{D}$.

Our IMTalker framework factorizes this problem into two main stages. 
First, the motion latent generation stage acquires the driving motion latent $z_{\text{motion, D}}$. This is either synthesized from audio $A$ (and optional control signals pose $p$, gaze $g$) by a flow matching motion generator ($G_{\text{motion}}$) for the audio-driven task, or directly extracted from a driving video $V_D$ by a motion encoder ($E_{\text{motion}}$) for the video-driven task. 
Second, the motion transfer and portrait rendering stage, the renderer $G_{\text{render}}$ synthesizes the final frame $\hat{I}_T$ conditioned on the driving motion $z_{\text{motion, D}}$ and two representations from the source image $I_S$: identity features $f_{id}$ (from $E_{\text{id}}$) and source motion $z_{\text{motion, S}}$ (from $E_{\text{motion}}$).
This complete process is formulated as:
\begin{gather}
f_{id}, z_{\text{motion, S}} = E_{\text{id}}(I_S), E_{\text{motion}}(I_S) 
\label{eq:source_extraction} \\[1.5mm]
z_{\text{motion, D}} = 
\begin{cases}
G_{\text{motion}}(A, p, g), & \text{if } \mathcal{D} \text{ is audio} \\
E_{\text{motion}}(V_D), & \text{if } \mathcal{D} \text{ is video}
\end{cases} 
\label{eq:driving_extraction} \\[1.5mm]
\hat{I}_T = G_{\text{render}}(f_{id}, z_{\text{motion, S}}, z_{\text{motion, D}})
\label{eq:render_updated}
\end{gather}

\subsection{Identity-Adaptive Motion Transfer Renderer}
\label{sec:render}
The Renderer $G_{\text{render}}$ is designed to follow a precise data flow and is composed of three key sub-modules: (1) an \textbf{Identity-Adaptive (IA) Module ($\Phi$)} that first personalizes the input motion latents (\cref{sec:IAD}), (2) an \textbf{Implicit Motion Transfer (IMT) Module} that then align motion discrepancies with dense identity features (\cref{sec:IMT}), and (3) a \textbf{Synthesis Network} that finally renders these aligned features into a photorealistic output image (\cref{sec:synthesis}). The training objective for this stage is described in \cref{sec:train1}.

\subsubsection{Identity-Adaptive (IA) Module}
\label{sec:IAD}
The core problem we target is that a single, global motion latent $z_{\text{motion}}$ is not optimal for all identities, as different individuals have unique ways of expressing the same phoneme or emotion. Therefore, we introduce a lightweight yet highly effective Identity-Adaptive  Module ($\Phi$). The function of $\Phi$ is to specialize this global motion vector for a target identity, embedding identity-specific style without corrupting the core motion command. 
$\Phi$ is a small MLP that takes the motion latent $z_{\text{motion}}$ and the global identity feature $f_{\text{global}}$ as input to produce a personalized motion latent $z'_{\text{motion}}$:
\begin{gather}
    \Phi: \mathbb{R}^{d_z} \times \mathbb{R}^{d_f} \to \mathbb{R}^{d_z} \\
    z'_{\text{motion}} = \Phi(z_{\text{motion}}, f_{\text{global}})
\end{gather}
However, in practical training, the source and driving frames originate from the same identity (see \cref{sec:train1}). This inevitably introduces an identity bias into the motion latent during our adaptive process, which in turn diminishes its ability to generalize to significant identity discrepancies. Therefore, we design a Motion Distance Consistency Loss $\mathcal{L}_{\text{dist}}$ to enforce the identity-invariance of the motion latent prior to identity adaptation. 
For each motion latent $z_{\text{motion}}$, we sample two distinct identities, $A$ and $B$. We then compute their respective personalized motion latents:
\begin{align}
    z'_{A} &= \Phi(z_{\text{motion}}, f_{\text{global}, A}) \\
    z'_{B} &= \Phi(z_{\text{motion}}, f_{\text{global}, B})
\end{align}
The loss then encourages the distance from the original motion latent to each of the personalized latents to be equal:
\begin{equation}
    \mathcal{L}_{\text{dist}} = \left| d(z_{\text{motion}}, z'_{A}) - d(z_{\text{motion}}, z'_{B}) \right|
    \label{eq:dist_loss}
\end{equation}
where $d(\cdot, \cdot)$ is the L1 distance. This objective forces all personalized motion latents to lie on a hypersphere centered at the original motion latent. It ensures that the adaptation module $\Phi$ applies a stylization of consistent ``strength" for all identities, preventing it from adapting some more aggressively than others. 

\subsubsection{Implicit Motion Transfer (IMT) Module}
\label{sec:IMT}
The goal of the Implicit Motion Transfer (IMT) Module is to perform motion transfer by generating new appearance features directly in the latent space, rather than simply rearranging existing ones. To initiate this, a unified motion decoder $D_{\text{motion}}$ utilizes style modulation \cite{karras2020analyzing} to transform the personalized latent vectors ($z'_{\text{motion, S}}, z'_{\text{motion, D}}$) into multi-scale 2D motion maps ($M_S, M_D$), where each scale corresponds to a different granularity of facial motion. These maps are spatially aligned with the dense identity features $f_{\text{dense}}$, establishing a crucial sub-pixel spatial correspondence between motion and appearance in a unified 2D latent space. At each scale, the 2D feature maps $M_D$, $M_S$, and $f_{\text{dense}}$ serve as the inputs for our cross-attention mechanism. However, the $O(N^2)$ quadratic complexity of standard attention imposes a prohibitive computational burden when processing high-resolution feature maps. To address this, we propose a hierarchical coarse-to-fine guided attention strategy. Our architecture processes coarse and fine feature maps differently:

At lower resolution (e.g., $64^2$), we employ a standard cross-attention mechanism. The input feature maps ($M_D, M_S, f_{\text{dense}}$) are linearly projected and flattened into sequences to form the Query ($Q$), Key ($K$), and Value ($V$) matrices. The full attention map, denoted as $A_{\text{coarse}}$, is then computed from $Q$ to $K$, and subsequently applied to $V$ to produce the aligned feature set $f_{\text{aligned, coarse}}$. 

At finer resolutions (e.g., $128^2$, $256^2$), we introduce a highly efficient Guided Sparse Resampler module that operates without computing QK attention. This module takes two inputs: the high-resolution value features $V_{\text{high}}$ (derived from $f_{\text{dense}}$) and the coarse attention map $A_{\text{coarse}}$ (output from the previous coarse stage). $A_{\text{coarse}}$ is first spatially upsampled to match the resolution of $V_{\text{high}}$. Then, for each row in the upsampled attention map, we select the top-$k$ values to form a sparse attention mask. This mask is subsequently applied to $V_{\text{high}}$ using the Hadamard product
 to generate the aligned feature set $f_{\text{aligned, fine}}$.

This efficient, hybrid strategy retains a global receptive field, overcoming the limitations of local affine approximations in traditional warp-based methods. It allows the resulting features to represent newly generated appearance, enabling the robust handling of complex, non-rigid facial motions. The set of multi-scale aligned features from all scales, denoted as $f_{\text{aligned}}$, is then passed to the Synthesis Network.
\subsubsection{Synthesis Network}
\label{sec:synthesis}
The Synthesis Network is responsible for fusing different aligned features $f_{\text{aligned}}$ from the IMT module into the final output image $\hat{I}_T$. Its architecture is composed of a stack of alternating transformer blocks and residual convolution (ResConv) blocks. This hybrid design is crucial for fusing global and local information at multiple scales. For efficiency considerations, we adopt standard Transformer blocks for $f_{\text{aligned, coarse}}$, and employ the shift windows mechanism similar to \cite{liu2021swin} for  $f_{\text{aligned, fine}}$ to avoid full-attention operations.

\subsubsection{Training Objective}
\label{sec:train1}
We jointly train the renderer $G_{\text{render}}$ and the encoders using a reconstruction objective in a self-supervised manner. Given a source frame $I_S$ and target frame $I_T$, we first extracts the source identity latent $f_id = E_id(I_S)$ and their respective motion latents, $z_S = E_{\text{motion}}(I_S)$ and $z_T = E_{\text{motion}}(I_T)$. The renderer then reconstructs the target frame as $\hat{I}_T = G_{\text{render}}(f_{id}, z_S, z_T)$. The networks are optimized by minimizing the following composite loss between $\hat{I}_T$ and $I_T$:
\begin{equation}
    \mathcal{L}_{\text{total}} = \lambda_{\text{rec}}\mathcal{L}_{\text{rec}} + \lambda_{\text{lpips}}\mathcal{L}_{\text{lpips}} + \lambda_{\text{GAN}}\mathcal{L}_{\text{GAN}}  + \lambda_{\text{dist}}\mathcal{L}_{\text{dist}}
    \label{eq:total_loss}
\end{equation}
The $\lambda$ terms balance the objectives for pixel-level reconstruction, perceptual similarity, photorealism, and representation disentanglement.

\begin{figure*}[t!]
    \centering
    \setlength{\tabcolsep}{1pt}
    \renewcommand{\arraystretch}{1.0}

    \begin{tabular}{cccccccc}
        \includegraphics[width=0.12\textwidth]{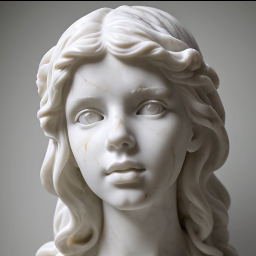} &
        \includegraphics[width=0.12\textwidth]{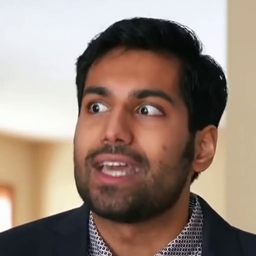} &
        \includegraphics[width=0.12\textwidth]{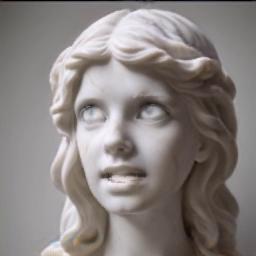} &
        \includegraphics[width=0.12\textwidth]{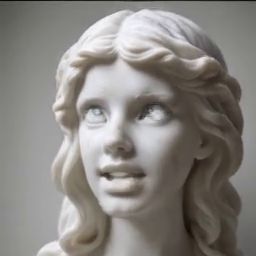} &
        \includegraphics[width=0.12\textwidth]{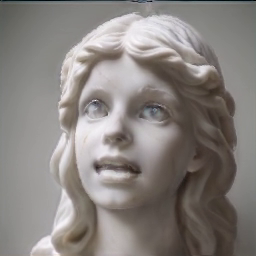} &
        \includegraphics[width=0.12\textwidth]{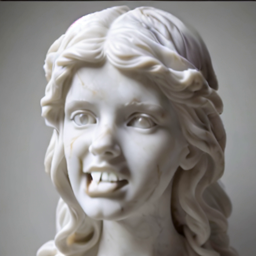} &
        \includegraphics[width=0.12\textwidth]{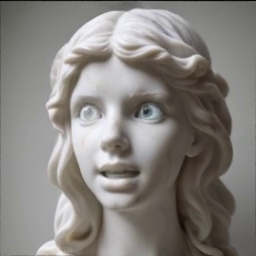} &
        \includegraphics[width=0.12\textwidth]{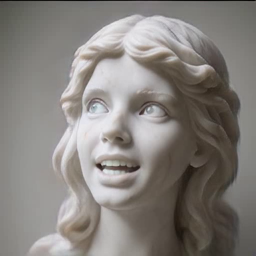} \\

        \includegraphics[width=0.12\textwidth]{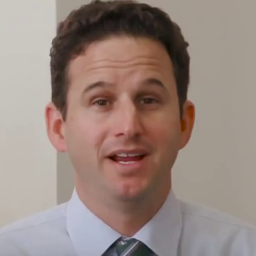} &
        \includegraphics[width=0.12\textwidth]{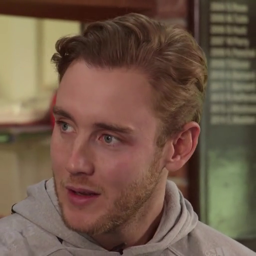} &
        \includegraphics[width=0.12\textwidth]{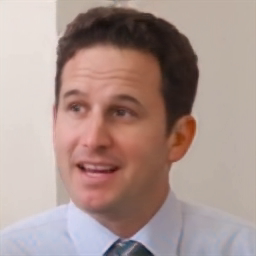} &
        \includegraphics[width=0.12\textwidth]{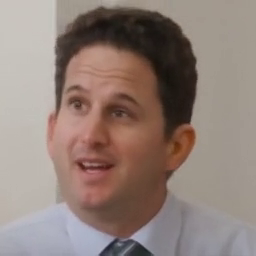} &
        \includegraphics[width=0.12\textwidth]{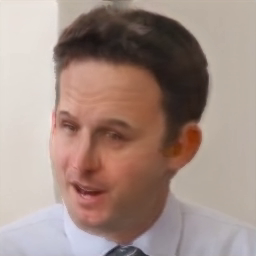} &
        \includegraphics[width=0.12\textwidth]{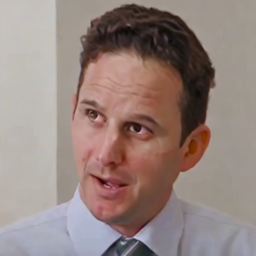} &
        \includegraphics[width=0.12\textwidth]{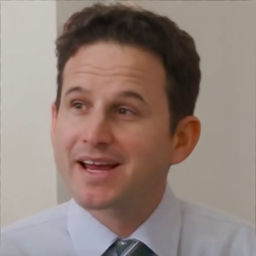} &
        \includegraphics[width=0.12\textwidth]{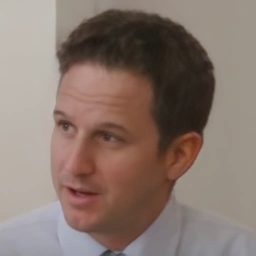} \\

        \includegraphics[width=0.12\textwidth]{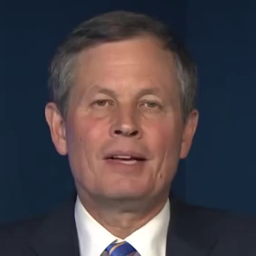} &
        \includegraphics[width=0.12\textwidth]{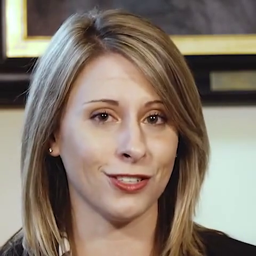} &
        \includegraphics[width=0.12\textwidth]{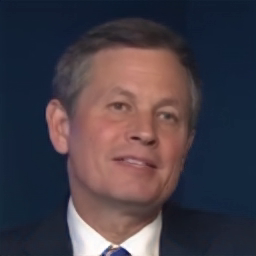} &
        \includegraphics[width=0.12\textwidth]{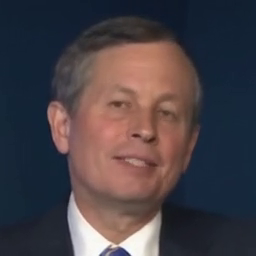} &
        \includegraphics[width=0.12\textwidth]{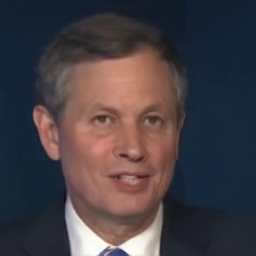} &
        \includegraphics[width=0.12\textwidth]{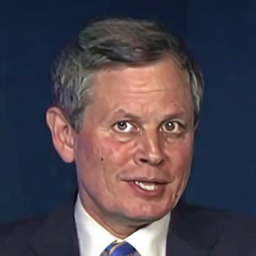} &
        \includegraphics[width=0.12\textwidth]{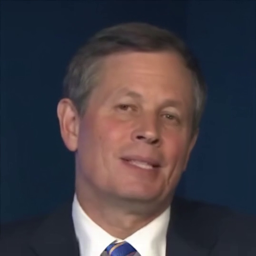} &
        \includegraphics[width=0.12\textwidth]{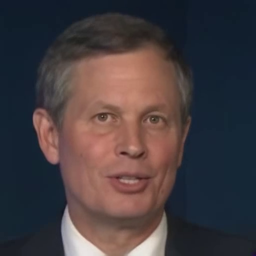} \\

        \includegraphics[width=0.12\textwidth]{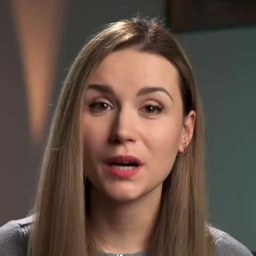} &
        \includegraphics[width=0.12\textwidth]{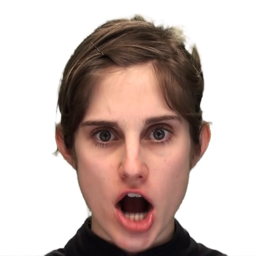} &
        \includegraphics[width=0.12\textwidth]{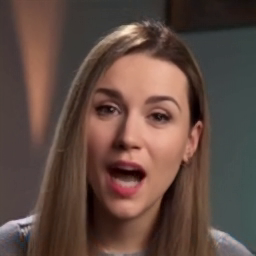} &
        \includegraphics[width=0.12\textwidth]{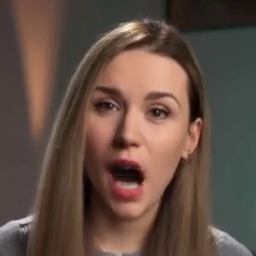} &
        \includegraphics[width=0.12\textwidth]{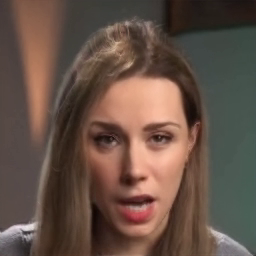} &
        \includegraphics[width=0.12\textwidth]{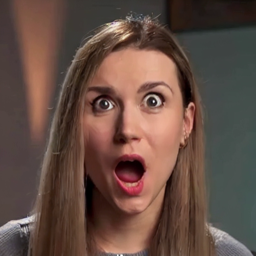} &
        \includegraphics[width=0.12\textwidth]{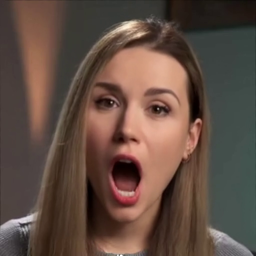} &
        \includegraphics[width=0.12\textwidth]{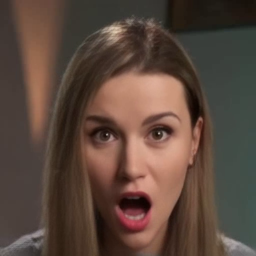} \\
        
        \textbf{Source} & \textbf{Driving} & LIA & MCNet & EDTalk & X-Portrait & LivePortrait & \textbf{IMTalker (Ours)}
    \end{tabular}

    \caption{
        \textbf{Qualitative comparison on cross-reenactment.}
        Our method (\textbf{IMTalker}) achieves superior identity preservation and visual realism compared to existing baselines.
        Please refer to our supplementary material for more detailed comparison.
    }
    \label{fig:video_driven_comparison}
\end{figure*}
\subsection{Flow Matching Motion Generator}
\label{sec:generator}
Inspired by recent two-stage models~\cite{liu2024anitalker, li2024ditto, ki2025float, ma2025playmate}, rather than inefficiently modeling the direct audio-to-pixel-space mapping, we design a lightweight flow matching based generator $G_{\text{motion}}$ to learn the mapping from audio signals to the low-dimensional motion latents (learned in the first stage). To further enhance the model's controllability, we also incorporate deterministic control conditions, such as pose and gaze, into the generator.

\subsubsection{Conditional Signals and Architecture}
We extract audio features ($f_{\text{audio}}$), 6D pose parameters ($p_{\text{6d}}$), and eye gaze ($g_{\text{eye}}$) from a pre-trained wav2vec2~\cite{baevski2020wav2vec} model, a 3D face reconstruction model~\cite{retsinas2024smirk}, and a gaze estimation model~\cite{abdelrahman2023l2cs}, respectively. These various conditions, along with the timestep $t$, are projected by their respective linear layers ($L_i$) and then fused via element-wise summation to form the final control condition $C$. We adopt a modified Diffusion Transformer (DiT) \cite{peebles2023scalable} as our motion generator backbone. Specifically, the DiT is composed of stacked self-attention layers and frame-wise Adaptive Layer Normalization~\cite{ba2016layer}(AdaLN). For an input feature map $h$ from a preceding layer (e.g., a self-attention block), the AdaLN layer uses $C$ to predict the scale ($\gamma_c$) and bias ($\beta_c$) parameters that modulate it:
\begin{equation}
    \text{AdaLN}(h) = \gamma_c \cdot \text{LayerNorm}(h) + \beta_c
\end{equation}
This design leverages self-attention to model temporal dependencies, while AdaLN provides precise, frame-level control over the synthesis process.

\subsubsection{Training Objective and Inference}
We parameterize our motion generator $G_{\text{motion}}$ as a vector field predictor $v_{\theta}$ and train it using a Flow Matching (FM) objective. The predictor $v_{\theta}$ learns to predict the straight-line vector field $z_1 - z_0$ (from a Gaussian prior $z_0$ to a ground-truth motion latent $z_1$), conditioned on the condition $C$ and timestep $t$. To enable Classifier-Free Guidance (CFG)~\cite{ho2022classifier}, during training, each control condition $m \in \{\text{audio}, \text{pose}, \text{gaze}\}$ is independently replaced by a null embedding $\emptyset_m$ with a probability of $p=0.1$. The FM objective is minimized via MSE:
\begin{equation}
    \mathcal{L}_{\text{FM}} = \mathbb{E}_{t, z_0, z_1, c} \left\| v_{\theta}( (1-t)z_0 + tz_1, C, t ) - (z_1 - z_0) \right\|_2^2
    \label{eq:fm_loss}
\end{equation}
At inference, a numerical solver (e.g., Euler) solves the probability flow ODE $\frac{dz}{dt} = v'_{\theta}(z, c, t)$ from $t=0$ to $t=1$. The guided vector field $v'_{\theta}$ is computed at each step using a guidance scale $w$:
\begin{equation}
    v'_{\theta}(z_t, c, t) = v_{\theta}(z_t, \emptyset, t) + w \cdot (v_{\theta}(z_t, c, t) - v_{\theta}(z_t, \emptyset, t))
    \label{eq:cfg}
\end{equation}
where $z_t$ is the current motion latent, and $v_{\theta}(z_t, \emptyset, t)$ is the  unconditional prediction, where $\emptyset = \{\emptyset_{\text{audio}}\}$.


%% file: sec/4_Experiment.tex
\begin{table*}[t] 
    \small 
    \centering
    \caption{Quantitative comparison for video-driven self-reenactment and cross-reenactment on the HDTF dataset. Best and second-best results are in \textbf{bold} and \underline{underlined}.}
    \label{tab:video_driven_combined} %
    \setlength{\tabcolsep}{5pt} %
    \renewcommand{\arraystretch}{1.05} %
    
    \begin{tabular}{@{}l
                    S[table-format=2.3] %
                    S[table-format=1.3]
                    S[table-format=1.3]
                    S[table-format=1.3]
                    S[table-format=2.3] %
                    @{\hskip 12pt} %
                    S[table-format=1.3]
                    S[table-format=1.3]
                    S[table-format=2.3] %
                    S[table-format=1.3] %
                    S[table-format=2.3] %
                    @{}}
        \toprule
        & \multicolumn{5}{c}{\textbf{Self-Reenactment}} %
        & \multicolumn{5}{c}{\textbf{Cross-Reenactment}} \\ %
        
        \cmidrule(lr){2-6} \cmidrule(lr){7-11} 
        
        \textbf{Method} & {\textbf{PSNR}$\uparrow$} & {\textbf{SSIM}$\uparrow$} & {\textbf{LPIPS}$\downarrow$} & {\textbf{CSIM}$\uparrow$} & {\textbf{FID}$\downarrow$} & {\textbf{AED}$\downarrow$} & {\textbf{APD}$\downarrow$} & {\textbf{MAE}$\downarrow$} & {\textbf{CSIM}$\uparrow$} & {\textbf{FID}$\downarrow$} \\ %
        
        \midrule
        LIA \cite{wang2022latent}          & 25.991 & 0.839 & 0.085 & 0.862 & 14.053 & 0.808 & 0.096 & 10.215 & 0.696 & 59.812 \\
        MCNet \cite{hong23implicit}        & \underline{27.974} & \underline{0.886} & 0.070 & 0.876 & 9.672 & 0.830 & 0.095 & 10.230 & 0.696 & 54.448 \\
        EDTalk \cite{tan2024edtalk}       & 26.366 & 0.853 & 0.081 & 0.868 & 12.201 & 0.713 & 0.038 & 10.583 & 0.750 & 60.669 \\
        X-Portrait \cite{xie2024x}   & 25.590 & 0.841 & 0.063 & 0.878 & 10.415 & 0.745 & 0.062 & 9.977 & 0.779 & 54.762 \\
        LivePortrait \cite{guo2024liveportrait} & 27.173 & 0.878 & \underline{0.043} & \underline{0.896} & \underline{9.049} & \underline{0.627} & \underline{0.026} & \underline{9.841} & \underline{0.789} & \underline{53.144} \\
        \midrule
        IMTalker (Ours) & \textbf{28.458} & \textbf{0.899} & \textbf{0.037} & \textbf{0.898} & \textbf{7.426} & \textbf{0.581} & \textbf{0.024} & \textbf{7.553} & \textbf{0.824} & \textbf{53.137} \\
        \bottomrule
    \end{tabular}
\end{table*}
\section{Experiment}
\label{sec:experiment}
We will first introduce our framework's implementation details, baselines, metrics, parameters and inference speed. Subsequently, we will present the experimental results under two different evaluation settings—video-driven and audio-driven—providing both qualitative and quantitative assessments. Finally, we conduct ablation studies to validate the effectiveness of our proposed modules.

\vspace{-8pt}
\paragraph{Implementation details}
For the Renderer, we trained the model on a composite dataset from three public talking head video datasets—VFHQ \cite{xie2022vfhq}, VoxCeleb2 \cite{chung2018voxceleb2}, and MultiTalk \cite{sung2024multitalk}—totaling approximately 660 hours (35,000 clips). Input frames were resized to $256^2$ resolution, with an output resolution of $512^2$. The motion generator utilized only the video and audio data from VoxCeleb2 \cite{chung2018voxceleb2}, comprising approximately 350 hours (160,000 clips). Both stages were trained on 4 NVIDIA A100 GPUs using the Adam optimizer \cite{kingma2014adam}, with a learning rate of $1 \times 10^{-4}$, $\beta_1 = 0.5$, and $\beta_2 = 0.99$. Using batch sizes of 16 and 256, the renderer and the generator were trained to full convergence in approximately 4 days and 2 days, respectively.
\begin{figure*}[t]
    \def\hdtf{sec/figs/audio_driven/hdtf}
    \def\celebv{sec/figs/audio_driven/celebv}
    
    \def\sourcehdtf{sec/figs/audio_driven/hdtf_source.png} 
    \def\sourcecelebv{sec/figs/audio_driven/celebv_source.png}

    \def\MyRowStrut{\vphantom{\includegraphics[width=0.085\linewidth]{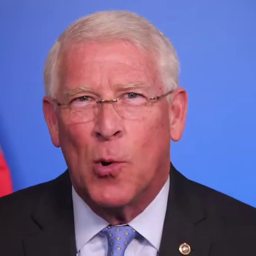}}}

    \resizebox{0.92\textwidth}{!}{
    \setlength{\tabcolsep}{1pt} 
    \renewcommand{\arraystretch}{0.95} 

    \begin{tabular}{l ccccc @{\hspace{8pt}} c @{\hspace{8pt}} ccccc}
        \textbf{Methods} &
        \multicolumn{5}{c}{\textbf{HDTF}} &
        \textbf{Source} &
        \multicolumn{5}{c}{\textbf{CelebV}} \\[3pt]
        
        \textbf{AniTalker} \MyRowStrut &
        \includegraphics[width=0.085\linewidth]{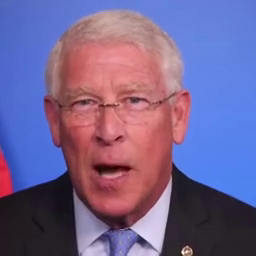} & 
        \includegraphics[width=0.085\linewidth]{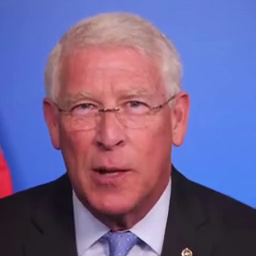} &
        \includegraphics[width=0.085\linewidth]{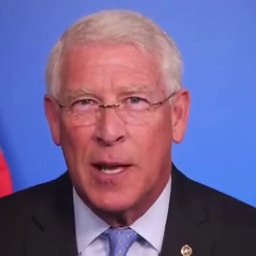} &
        \includegraphics[width=0.085\linewidth]{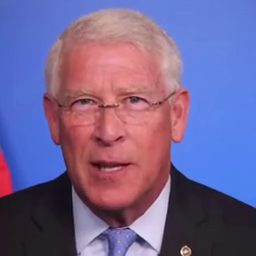} &
        \includegraphics[width=0.085\linewidth]{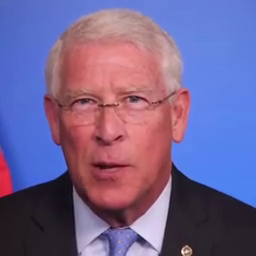} &
        & 
        \includegraphics[width=0.085\linewidth]{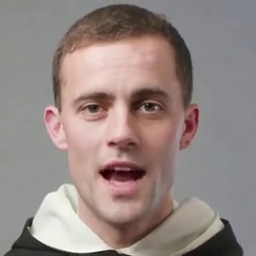} &
        \includegraphics[width=0.085\linewidth]{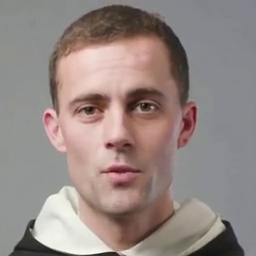} &
        \includegraphics[width=0.085\linewidth]{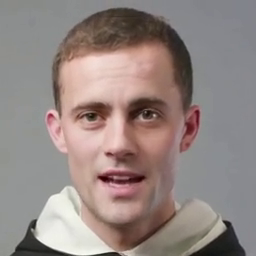} &
        \includegraphics[width=0.085\linewidth]{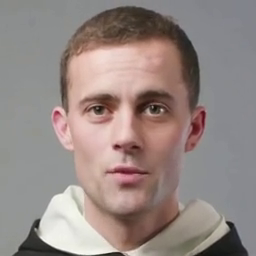} &
        \includegraphics[width=0.085\linewidth]{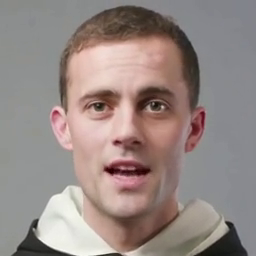} \\

        \textbf{Hallo} \MyRowStrut &
        \includegraphics[width=0.085\linewidth]{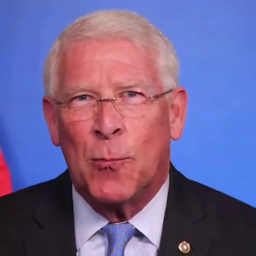} &
        \includegraphics[width=0.085\linewidth]{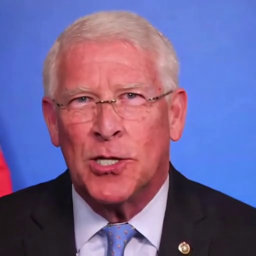} &
        \includegraphics[width=0.085\linewidth]{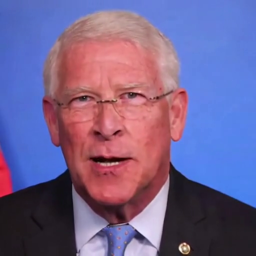} &
        \includegraphics[width=0.085\linewidth]{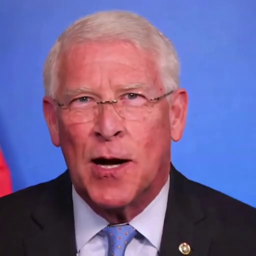} &
        \includegraphics[width=0.085\linewidth]{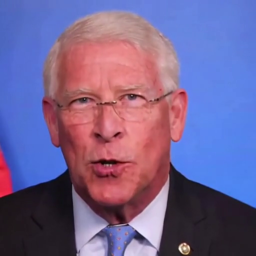} &
        \includegraphics[width=0.085\linewidth]{\sourcehdtf} & 
        \includegraphics[width=0.085\linewidth]{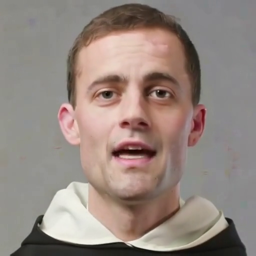} &
        \includegraphics[width=0.085\linewidth]{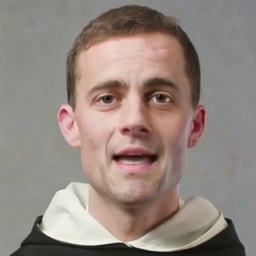} &
        \includegraphics[width=0.085\linewidth]{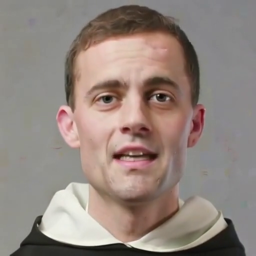} &
        \includegraphics[width=0.085\linewidth]{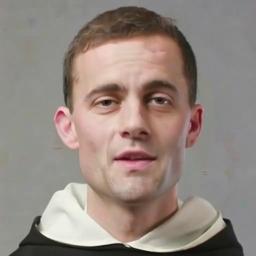} &
        \includegraphics[width=0.085\linewidth]{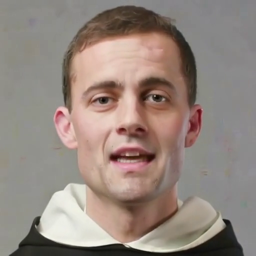} \\

        \textbf{EchoMimic} \MyRowStrut &
        \includegraphics[width=0.085\linewidth]{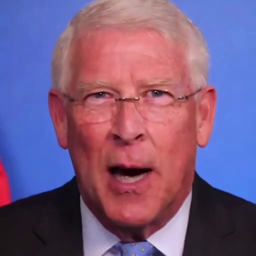} &
        \includegraphics[width=0.085\linewidth]{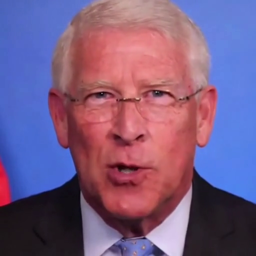} &
        \includegraphics[width=0.085\linewidth]{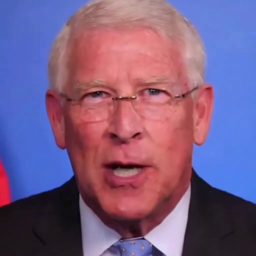} &
        \includegraphics[width=0.085\linewidth]{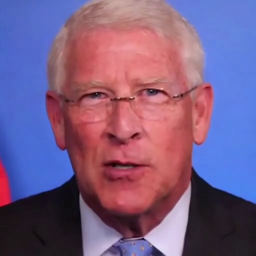} &
        \includegraphics[width=0.085\linewidth]{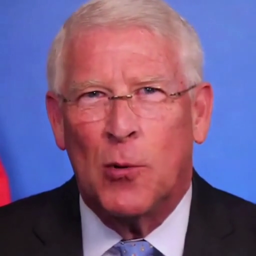} &
        & 
        \includegraphics[width=0.085\linewidth]{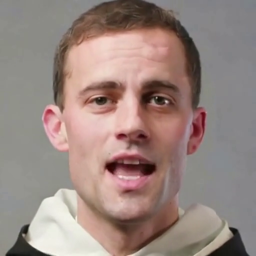} &
        \includegraphics[width=0.085\linewidth]{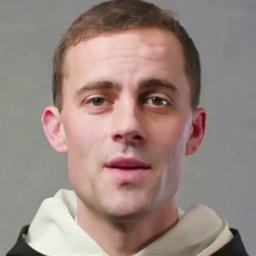} &
        \includegraphics[width=0.085\linewidth]{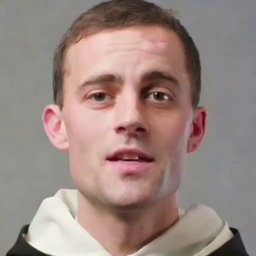} &
        \includegraphics[width=0.085\linewidth]{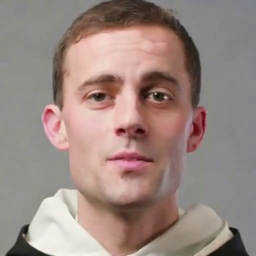} &
        \includegraphics[width=0.085\linewidth]{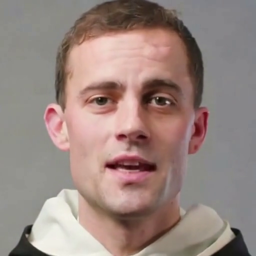} \\

        \textbf{Ditto} \MyRowStrut &
        \includegraphics[width=0.085\linewidth]{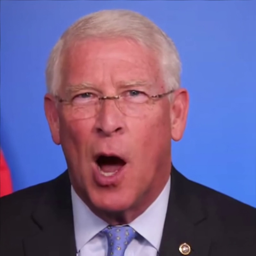} &
        \includegraphics[width=0.085\linewidth]{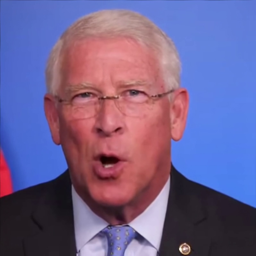} &
        \includegraphics[width=0.085\linewidth]{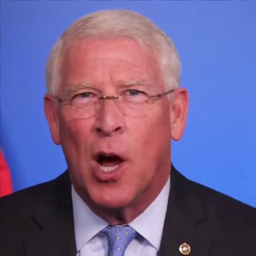} &
        \includegraphics[width=0.085\linewidth]{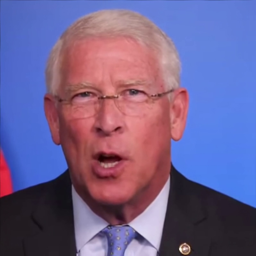} &
        \includegraphics[width=0.085\linewidth]{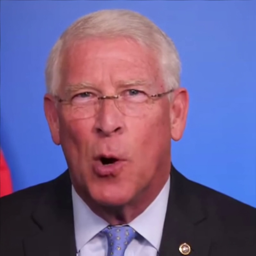} &
        & 
        \includegraphics[width=0.085\linewidth]{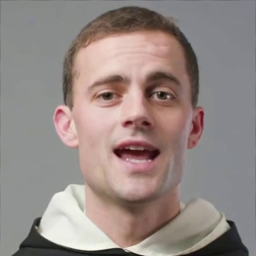} &
        \includegraphics[width=0.085\linewidth]{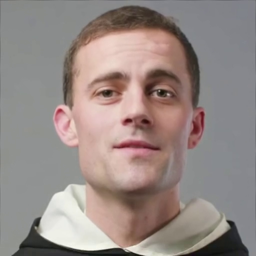} &
        \includegraphics[width=0.085\linewidth]{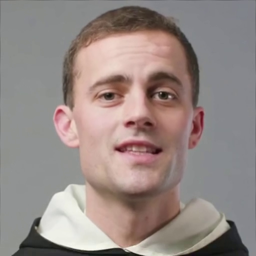} &
        \includegraphics[width=0.085\linewidth]{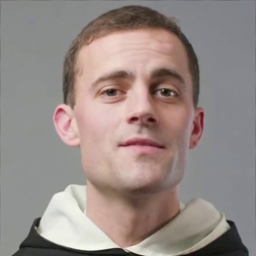} &
        \includegraphics[width=0.085\linewidth]{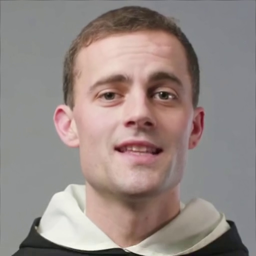} \\

        \textbf{FLOAT} \MyRowStrut &
        \includegraphics[width=0.085\linewidth]{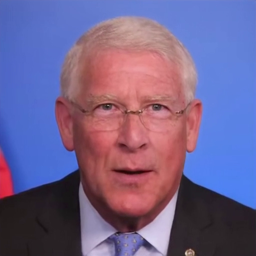} &
        \includegraphics[width=0.085\linewidth]{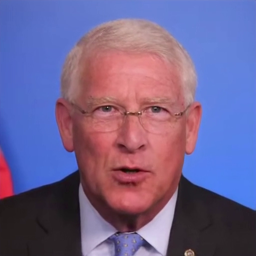} &
        \includegraphics[width=0.085\linewidth]{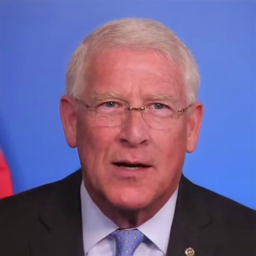} &
        \includegraphics[width=0.085\linewidth]{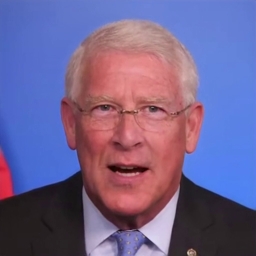} &
        \includegraphics[width=0.085\linewidth]{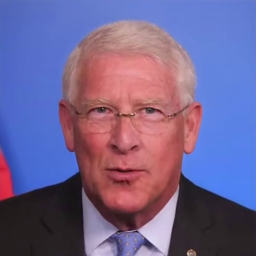} &
        \includegraphics[width=0.085\linewidth]{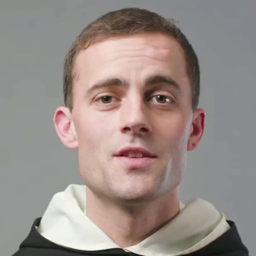} & 
        \includegraphics[width=0.085\linewidth]{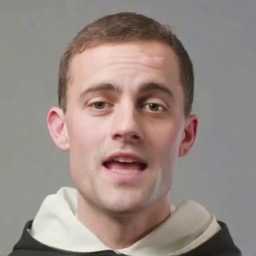} &
        \includegraphics[width=0.085\linewidth]{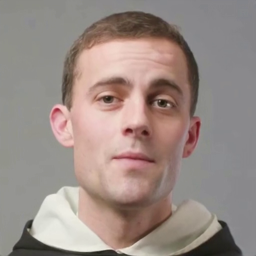} &
        \includegraphics[width=0.085\linewidth]{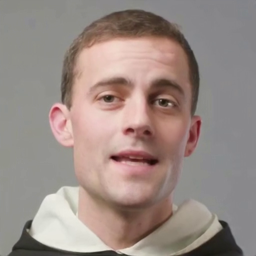} &
        \includegraphics[width=0.085\linewidth]{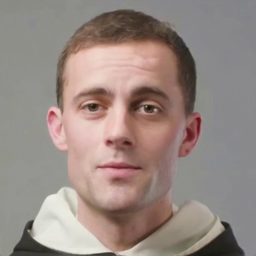} &
        \includegraphics[width=0.085\linewidth]{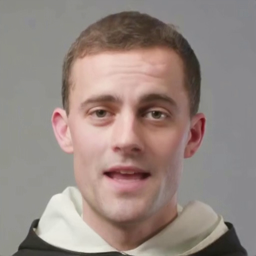} \\

        \textbf{IMTalker} \MyRowStrut &
        \includegraphics[width=0.085\linewidth]{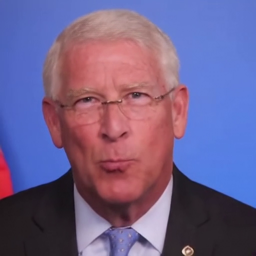} &
        \includegraphics[width=0.085\linewidth]{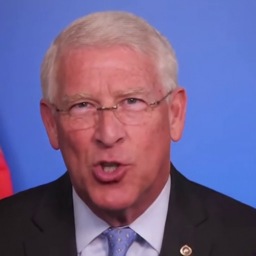} &
        \includegraphics[width=0.085\linewidth]{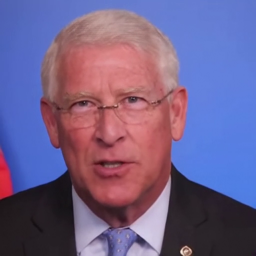} &
        \includegraphics[width=0.085\linewidth]{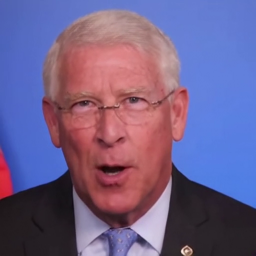} &
        \includegraphics[width=0.085\linewidth]{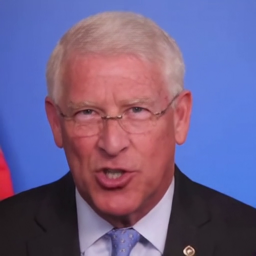} &
        &
        \includegraphics[width=0.085\linewidth]{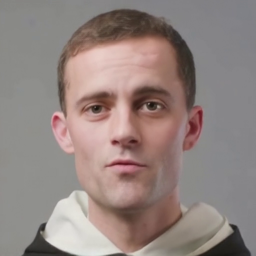} &
        \includegraphics[width=0.085\linewidth]{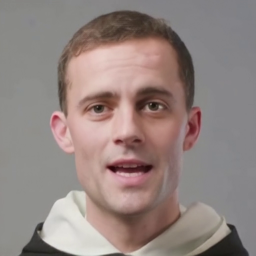} &
        \includegraphics[width=0.085\linewidth]{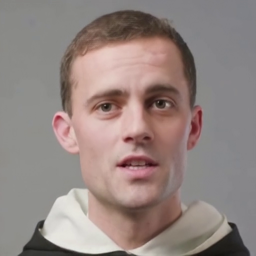} &
        \includegraphics[width=0.085\linewidth]{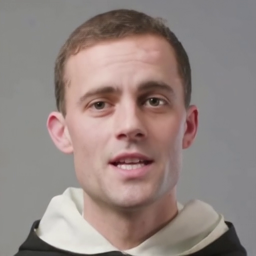} &
        \includegraphics[width=0.085\linewidth]{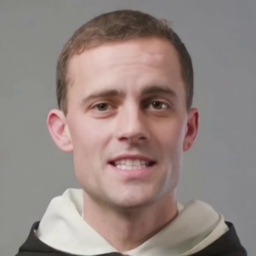} \\
    \end{tabular}%
    } 

    \vspace{1pt}
    \caption{
        Qualitative comparison of audio-driven talking head generation results. 
        The \textbf{Source} column (center) shows the single source identity used for all methods on \textbf{HDTF} \cite{zhang2021flow} (top image, left block) and \textbf{CelebV} \cite{yu2023celebv} (bottom image, right block).
        Each row corresponds to one method's results on different test samples.
        Please refer to our supplementary material for more detailed comparison.
    }
    \label{fig:audio_driven_compare}
    \vspace{-8pt}
\end{figure*}

\vspace{-6pt}
\paragraph{Baselines}
In the video-driven setting, we compare our method against several open-source baselines, including LIA \cite{wang2022latent}, MCNet \cite{hong2023implicit}, EDTalk \cite{tan2024edtalk}, LivePortrait \cite{guo2024liveportrait}, and X-Portrait \cite{xie2024x}. In the audio-driven setting, we conduct comparisons with EchoMimic \cite{chen2025echomimic}, Hallo \cite{xu2024hallo},  AniTalker \cite{liu2024anitalker}, Ditto \cite{li2024ditto}, and FLOAT \cite{ki2025float}.

\paragraph{Metrics}
We use Peak Signal-to-Noise Ratio (PSNR),  Structural Similarity Index (SSIM) \cite{wang2004image}, Learned Perceptual Image Patch Similarity (LPIPS) \cite{zhang2018unreasonable},  Fréchet Inception Distance (FID) \cite{heusel2017gans}, and Fréchet Video Distance (FVD) \cite{unterthiner2018towards} to evaluate the image and video generation quality. To assess identity consistency and motion transfer accuracy, we employ Cosine Similarity of identity embedding (CSIM) \cite{deng2019arcface}, Average Expression Distance (AED) \cite{siarohin2019first}, Average Pose Distance (APD) \cite{siarohin2019first}, and  Mean Angular Error (MAE) \cite{han2024face} of eyeball direction. The degree of audio-lip synchronization is measured using  Lip-Sync Error Confidence (Sync-C) \cite{prajwal2020lip} and Lip-Sync Error Distance (Sync-D) \cite{prajwal2020lip}. 
\begin{table*}[t]
    \centering
    \small 
    \setlength{\tabcolsep}{5pt} 
    \renewcommand{\arraystretch}{1} 
    \caption{
        Comparison of audio-driven methods on the HDTF and CelebV datasets. Best and second-best results are in \textbf{bold} and \underline{underlined}.
    }
    \label{tab:audio_driven}
    \begin{tabular}{
        @{}l
        S[table-format=2.3]@{/}S[table-format=2.3]
        S[table-format=3.3]@{/}S[table-format=3.3]
        S[table-format=1.3]@{/}S[table-format=1.3]
        S[table-format=1.3]@{/}S[table-format=1.3]
        S[table-format=1.3]@{/}S[table-format=1.3]
        S[table-format=1.3]@{/}S[table-format=1.3]
        S[table-format=1.3]@{/}S[table-format=1.3]
        @{}}
        \toprule
        \textbf{Method} & \multicolumn{2}{c}{\textbf{FID}$\downarrow$} & \multicolumn{2}{c}{\textbf{FVD}$\downarrow$} & \multicolumn{2}{c}{\textbf{Sync-C}$\uparrow$} & \multicolumn{2}{c}{\textbf{Sync-D}$\downarrow$} & \multicolumn{2}{c}{\textbf{CSIM}$\uparrow$} & \multicolumn{2}{c}{\textbf{SSIM}$\uparrow$} & \multicolumn{2}{c}{\textbf{LPIPS}$\downarrow$} \\
        \midrule
        EchoMimic \cite{chen2025echomimic}   & 27.846 & 33.243 & 288.457 & 299.913 & 6.265 & 6.014 & 8.661 & 8.754 & 0.806 & 0.611 & 0.493 & 0.482 & 0.249 & 0.303 \\
        Hallo \cite{xu2024hallo}      & 10.573 & 21.538 & \underline{154.975} & 246.266 & \underline{7.497} & \underline{6.843} & \textbf{7.694} & \underline{8.065} & 0.863 & 0.787 & \underline{0.668} & 0.575 & \textbf{0.137} & \underline{0.233} \\
        Anitalker \cite{liu2024anitalker}  & 14.224 & 24.815 & 241.968 & 241.968 & 6.855 & 6.501 & 8.184 & 8.316 & 0.857 & 0.746 & 0.607 & 0.544 & 0.195 & 0.278 \\
        Ditto \cite{li2024ditto}      & 11.746 & 25.367 & 195.848 & 288.804 & 6.093 & 5.154 & 8.938 & 9.771 & \textbf{0.886} & \underline{0.806} & \textbf{0.669} & \underline{0.582} & 0.139 & 0.234 \\
        FLOAT \cite{ki2025float}      & \underline{9.164} & \underline{18.272} & 198.964 & \underline{228.215} & 6.854 & 6.843 & 8.336 & 8.171 & 0.843 & 0.745 & 0.639 & 0.577 & 0.161 & 0.239 \\
        \midrule
        IMTalker (Ours) & \textbf{9.084} & \textbf{17.921} & \textbf{143.623} & \textbf{200.592} & \textbf{7.711} & \textbf{7.364} & \underline{7.794} & \textbf{7.832} & \underline{0.869} & \textbf{0.814} & 0.665 & \textbf{0.585} & \underline{0.141} & \textbf{0.226} \\
        \bottomrule
    \end{tabular}
    \vspace{-8pt}
\end{table*}

\vspace{-5pt}
\paragraph{Parameters and Inference Speed}
Our renderer and motion generator comprise 124M and 39M learnable parameters, respectively. On a NVIDIA RTX 4090 GPU, they can generate $512^2$ resolution video at 40 FPS and 42 FPS, respectively, demonstrating that our method is sufficiently lightweight and fast for real-time applications.
\subsection{Video-Driven Setting}
In this video-driven setting, we compare the capabilities of our method against state-of-the-art (SOTA) baselines in terms of motion transfer, identity preservation, and image quality. Specifically, a source frame and a driving frame are provided as input, supplying the identity and motion information, respectively. Under this setting, if the source frame and driving frame originate from the same video, we term this self-reenactment; if they come from different videos, it is termed cross-reenactment. For a fair comparison, we cropped the output videos from all methods to $256^2$ resolution. The detailed quantitative and qualitative evaluations are as follows.

\vspace{-5pt}
\paragraph{Qualitative Evaluation}
For the qualitative comparison, to more intuitively demonstrate the superiority of our method in motion transfer across diverse subjects and motions, we sampled a wider variety of frames from datasets including CelebV \cite{yu2023celebv}, VFHQ \cite{xie2022vfhq}, and RAVDESS \cite{livingstone2018ryerson}. \cref{fig:video_driven_comparison} illustrates the qualitative evaluation results for cross-reenactment against other methods. Each row highlights specific challenging scenarios, such as eye gaze, large-angle poses, cross-gender reenactment, and rich emotional expressions. The baseline methods inevitably suffer from failure cases, including identity degradation and inaccurate motion transfer. In contrast, our method successfully handles these challenges and achieves significantly better results.
\vspace{-5pt}
\paragraph{Quantitative Evaluation}
For quantitative evaluation, we randomly sampled 50 videos from the HDTF \cite{zhang2021flow} dataset. The results are presented in \cref{tab:video_driven_combined}. 
In the self-reenactment setting (left), our method achieves the best results across all pixel-level and image quality metrics, indicating superior generation quality. 
In the cross-reenactment setting (right), IMTalker again achieves state-of-the-art results, showing particular strength in metrics like CSIM, AED, and MAE. This confirms our method's ability to accurately transfer fine-grained motions, such as expressions and eye gaze, while maintaining high identity consistency.
\subsection{Audio-driven Setting}
We use the HDTF \cite{zhang2021flow} and CelebV \cite{yu2023celebv} datasets for qualitative and quantitative comparisons against the baselines. Specifically, we randomly sampled 50 videos from each of these two datasets. For each video, we use the first frame as the source frame and the complete audio stream as the driving condition to generate the output video. The qualitative and quantitative evaluations are as follows.
\vspace{-5pt}
\paragraph{Qualitative Evaluation}
\cref{fig:audio_driven_compare} presents the qualitative comparison for the audio-driven task. As shown, other methods either fail to generate rich head movements and expressions, struggle to align fine-grained syllables with corresponding lip motions, or suffer from artifacts in high-frequency details (such as distorted teeth). In contrast, our method simultaneously achieves accurate lip synchronization, rich expressions and movements, and high-quality talking head video generation.
\vspace{-5pt}
\paragraph{Quantitative Evaluation}
We present the quantitative comparison results in \cref{tab:audio_driven}. As shown in the table, our method achieves the best performance on both datasets across most metrics, with the exception of CSIM, where it is slightly lower than Ditto. This discrepancy is attributable to the fact that Ditto generates only very slight head movements, and the CSIM metric is highly sensitive to pose variations. In contrast, our method not only achieves accurate lip synchronization but also generates rich and diverse head movements, which contributes to a higher degree of realism.
\begin{figure}[t]
    \centering
    \setlength{\tabcolsep}{1pt}
    \resizebox{0.95\columnwidth}{!}{
    \renewcommand{\arraystretch}{1.0}

    \begin{tabular}{cccc}
        \begin{minipage}[c]{0.22\linewidth}
            \centering
            \vspace{5pt}
            \includegraphics[height=1.8cm]{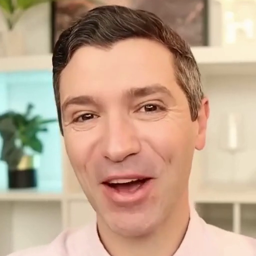}
            \vspace{5pt}
        \end{minipage} &
        \begin{minipage}[c]{0.22\linewidth}
            \centering
            \includegraphics[height=1.8cm]{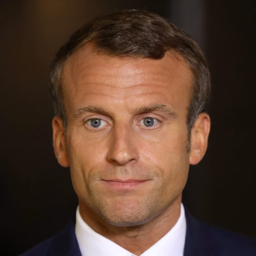}\\
            \includegraphics[height=1.8cm]{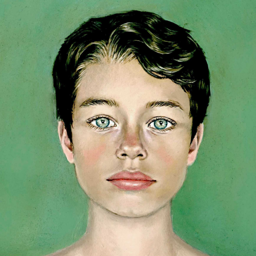}
        \end{minipage} &
        \begin{minipage}[c]{0.22\linewidth}
            \centering
            \includegraphics[height=1.8cm]{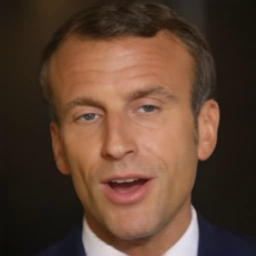}\\
            \includegraphics[height=1.8cm]{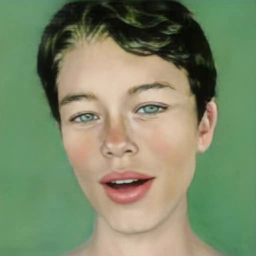}
        \end{minipage} &
        \begin{minipage}[c]{0.22\linewidth}
            \centering
            \includegraphics[height=1.8cm]{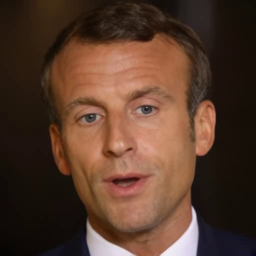}\\
            \includegraphics[height=1.8cm]{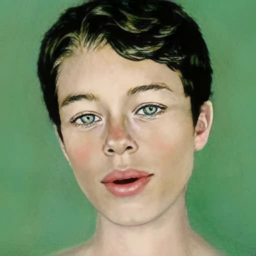}
        \end{minipage} \\[3pt]
        \textbf{Driving} & \textbf{Source} & \textbf{w/o IA} & \textbf{w IA} \\
    \end{tabular}
    }

    \vspace{-4pt}
    \caption{
        \textbf{Ablation study on Identity-Adaptive module}
        From left to right: driving frame, source images, results of our model without identity-adaptive module, and the full model results.
    }
    \vspace{-4pt}
    \label{fig:ablation on IAD}
\end{figure}
\begin{table}[t]
    \centering
    \caption{Ablation study on sampling steps and conditional inputs.}
    \label{tab:ablation_steps_pose_gaze}
    
    \resizebox{0.95\columnwidth}{!}{%
    \setlength{\tabcolsep}{3pt} 
    \renewcommand{\arraystretch}{1.0} 
    \begin{tabular}{@{}llcccc@{}}
        \toprule
        \multicolumn{2}{c}{\textbf{Setting}} & \textbf{FID}$\downarrow$ & \textbf{FVD}$\downarrow$ & \textbf{Sync-C}$\uparrow$ & \textbf{Sync-D}$\downarrow$ \\
        \midrule
        \multirow{3}{*}{\parbox[c]{1.55cm}{\centering \textit{No Pose}\\\textit{No Gaze}}} 
            & steps=5  & 17.97 & 206.32 & 7.43 & 7.65 \\
            & steps=10 & 17.55 & 200.96 & 7.36 & 7.83 \\
            & steps=50 & 17.49 & 200.59 & 7.24 & 7.98 \\
        \midrule
        \multirow{3}{*}{\centering \textit{steps=10}} 
            & +Pose        &  9.73 & 118.59 & 7.17 & 7.91 \\
            & +Gaze        & 16.11 & 191.42 & 7.26 & 7.89 \\
            & +Pose, Gaze  &  9.08 & 108.71 & 7.12 & 7.96 \\
        \bottomrule
    \end{tabular}%
    } 
\end{table}
\subsection{Ablation Study}
\paragraph{Ablation on Identity-Adaptive Module}
We conducted an ablation study by comparing the model with and without the Identity-Adaptive (IA) Module in the cross-reenactment setting. As visualized in \cref{fig:ablation on IAD}, without IA, the source identity's appearance is significantly degraded during the motion transfer process. This result demonstrates that IA disentangles identity content from motion and enables personalized motion adaptation, thereby significantly improving both identity consistency and motion accuracy.
\vspace{-5pt}
\paragraph{Ablation on Sampling Steps and Conditional Inputs}
We evaluate the impact of sampling steps and control conditions in \cref{tab:ablation_steps_pose_gaze}. In the audio-only setting, fewer sampling steps yield superior lip synchronization while maintaining stable video quality, highlighting the efficiency of our flow matching process. Furthermore, incorporating additional control conditions significantly enhances visual fidelity with negligible impact on lip alignment, demonstrating the robust controllability of our approach.

%% file: sec/5_Conclusion.tex
\vspace{-5pt}
\section{Conclusion}
\label{sec:conclusion}
We presented IMTalker, an efficient talking face generation framework driven by video or audio. Our core innovation is an Identity-Adaptive Motion Transfer Renderer that leverages cross-attention to perform motion transfer while disentangling identity-specific styles. For the audio-driven task, a lightweight Flow-Matching Motion Generator efficiently synthesizes controllable motion latents from multi-modal conditions. Extensive experiments on various benchmarks confirm that IMTalker achieves state-of-the-art performance in motion accuracy, identity preservation, and audio-lip synchronization, facilitating real-world applications.